\pdfoutput=1
\documentclass[11pt]{article}
\usepackage[hyperref]{emnlp2021}
\usepackage{times}
\usepackage{latexsym}
\usepackage[T1]{fontenc}
\usepackage[utf8]{inputenc}

\usepackage{microtype}

\usepackage{linguex}
\usepackage{subfig}
\usepackage{tikz}
\usetikzlibrary{shapes.geometric,backgrounds}

\usepackage{microtype}

\title{Transformers in the loop: \\  Polarity in neural models of language}

\author{Lisa Bylinina\Thanks{     Equal contribution. Accepted to ACL main conference. } \\
  Bookarang, Amsterdam  \\
  \texttt{bylinina@gmail.com} \\\And
  Alexey Tikhonov\footnotemark[1] \\
  Yandex Technologies GmbH, Berlin \\
  \texttt{altsoph@gmail.com} \\}

\date{}

\begin{document}
\maketitle
\begin{abstract}

Representation of linguistic phenomena in computational language models is typically assessed against the predictions of existing linguistic theories of these phenomena. Using the notion of polarity as a case study, we show that this is not always the most adequate set-up. We probe polarity via so-called `negative polarity items' (in particular, English {\it any}) in two pre-trained Transformer-based models (BERT and GPT-2). We show that -- at least for polarity -- metrics derived from language models are more consistent with data from psycholinguistic experiments than  linguistic theory predictions. Establishing this allows us to more adequately evaluate the performance of language models and also to use language models to discover new insights into natural language grammar beyond existing linguistic theories. This work contributes to establishing closer ties between psycholinguistic experiments and experiments with language models. 
\end{abstract}

\section{Introduction}

Recent Transformer-based language representation models (LRMs) -- such as BERT and GPT-2 \citep{bert,gpt} -- show impressive results on practical text analysis tasks. But do these models have access to complex linguistic notions? 
The results in this domain are less clear -- as well as ways to best approach this question.

Instead of asking whether LRMs encode fragments of current linguistic theory, we will directly compare metrics derived from LRMs to corresponding human judgments obtained in psycholinguistic experiments. The motivation for this is twofold. First, linguistic theories can be inaccurate -- so, evaluating a model with respect to predictions of such theories is not informative about the model performance. Second, robust abstract theoretical notions rarely correspond to robust judgments in humans, and `theoretical' and `perceived' versions of the same phenomenon can be significantly different (for instance, see \citealt{geurts} on inference judgments; discussed in Section 2). If this is something that LRMs inherit through training on human-produced texts, this makes LRMs an attractive possible component in an experimental pipeline, serving as a source of empirical predictions about human linguistic behaviour \citep{baroni2021,linzen2021}. 

As a case study, we focus on {\bf polarity}: a complex property of sentences at the intersection of grammar and semantics. We tackle polarity via the distribution of items that are sensitive to it -- namely, so-called {\bf negative polarity items} (NPIs) like English {\it any}. As a basic illustration of NPI sensitivity to polarity, consider a pair of sentences in \Next (* = ungrammaticality):

\vspace*{-1ex}
\ex. \a. Mary didn't buy any books.
\b. *Mary bought any books.

\vspace*{-1ex}
\Last[a] is a negative sentence (has negative polarity), and {\it any} is grammatical in it. \Last[b] is an affirmative sentence (has positive polarity) and {\it any} in this sentence is grammatically degraded compared to \Last[a]. Apart from this paradigmatic contrast, as we discuss below, polarity contrasts are expressed in a variety of ways and are tied to semantics.

As a proxy for a grammaticality measure, we will use the probability of {\tt any} in the masked token position (in BERT) (following \citealt{goldberg,npi} a.o.) and perplexity increase when adding {\tt any} to a sentence (in GPT-2). The differences in the metrics for the two different models stem from the differences in their architecture and training objectives. For all experiments, we use non-fine-tuned pre-trained LRMs. For this, we introduce our {\sc any} dataset, which combines natural and synthetic data.

{\bf We find} high levels of alignment between results of psycholinguistic experiments on monotonicity and NPIs, on the one hand -- and our LRM-derived results, on the other hand. Furthermore, show how LRMs can be used to make new predictions about NPIs in contexts with different numerals and confirm these predictions in a psycholinguistic experiment.

This case study contributes to the complement of the `interpretability of neural LRMs' research agenda: we can ask not only what linguistic tasks tell us about LRMs, but also what these models can help us find out about natural language (see \citealt{baroni2021,linzen2021} for a discussion along these lines). 

The paper is structured as follows. First, in section 2, we set up the context for our study: we describe the background in theoretical and experimental linguistics in the domains relevant for our discussion. Section 3 discusses previous work on NPIs and polarity in computational linguistics. Section 4 contains the description of our experimental method. First, we introduce our {\sc any} dataset; then, we describe the tests and metrics we use with BERT and with GPT-2 given our dataset. Section 5 discusses our results. In section 6, we go beyond state-of-the-art knowledge in experimental semantics and pragmatics and study the effect of the numeral on NPI acceptability -- first, we do a BERT study and then confirm the results on human participants. Section 7 concludes: we propose directions for future work aligning experimental studies of language in humans and LRMs.

\section{Background}

NPIs are expressions with limited linguistic distribution. While their use is grammatical in some sentences, in other sentences their use results in ungrammaticality. The distribution of NPIs like {\it any} is governed by the notion of polarity that is much more intricate than the simple presence or absence of sentential negation, as in \Last. 

For instance, in examples \Next-\NNext, \Next are `negative enough' to allow for (=`license') {\it any}, while \NNext are not -- even though none of these sentences contain overt sentential negation.

\vspace*{-1ex}
\ex. \a. None of the boxes contain anything.
\b. Nobody talked to anybody.
\b. At most five students did anything.
\b. Few people had any thoughts

\vspace*{-3ex}
\ex. \a. *Some of the boxes contain anything.
\b. *Somebody talked to anybody.
\b. *At least 5 students did anything.
\b. *Many people had any thoughts

\vspace*{-1ex}
The notion of polarity at play here relates to a semantic notion of {\bf monotonicity}.\footnote{This is a simplification. This is true of so-called `weak NPIs' -- a subclass of NPIs to which {\it any} belongs. We will keep referring to them simply as NPIs since we are only discussing weak ones. There are also other factors in weak NPI distribution apart from monotonicity (see \citealt{anastasia,barker}). Still, we focus on monotonicity as a crucial factor in NPI acceptability, following evidence discussed in the rest of the section.}

The notion of {\bf monotonicity} builds on logical entailment. Monotonicity of a linguistic environment defines its entailment patterns. In \Next, the domain in square brackets is {\bf upward-entailing} (UE), or upward-monotone, -- as evidenced by the valid inference from sets ({\it textbooks}) to supersets ({\it books}): sentence \Next[b] entails sentence \Next[a]. 

\vspace*{-1ex}
\ex.\label{ue} \a. Some boxes [ contain books ]$_{\uparrow}$
\b. Some boxes [ contain textbooks ]$_{\uparrow}$

\vspace*{-1ex}
In contrast, \Next shows a {\bf downward-entailing} (DE), or downward-monotone, environment, which supports inferences from sets ({\it books}) to subsets ({\it textbooks}): \Next[a] entails \Next[b].

\vspace*{-1ex}
\ex.\label{de} \a. No boxes [ contain books ]$_{\downarrow}$
\b. No boxes [ contain textbooks ]$_{\downarrow}$

\vspace*{-1ex}
Not all environments are either UE or DE -- some are {\bf non-monotone}, that is, supporting neither of the inferences:

\vspace*{-1ex}
\ex. \a. Exactly 5 boxes [ contain books ]$_-$
\b. Exactly 5 boxes [ contain textbooks ]$_-$

\vspace*{-1ex}
Expressions responsible for monotonicity of a linguistic context are a heterogeneous class that includes sentential operators such as negation and conditional {\it if}; quantifiers ({\it some}, {\it no}, {\it few}, {\it at most five} etc.); quantificational adverbs ({\it rarely}, {\it always} etc.) and more.

Monotonicity is a highly abstract logical property interfacing with general reasoning. At the same time, it is deeply embedded into natural language grammar and it is relevant for understanding of inner workings of different linguistic expressions, such as NPIs.

As shown by examples (1)-(3), DE contexts give rise to negative polarity, as seen from NPI acceptability; UE contexts are positive. There is conflicting evidence concerning non-monotone contexts \citep{crnic,sub2019}.

\begin{table*}[t]
  \centering
  \begin{tabular}{ l l || l l }
  \hline
    \multicolumn{2}{c}{Logical monotonicity} & \multicolumn{2}{c}{Subjective monotonicity}  \\
    \hline
{{\sc neg} $>>$ {\sc aff};} & {{\sc at most} $>$ {\sc at least}} & {{\sc neg} $>$ {\sc at most};} & {{\sc no} $>$ {\sc few}}\\
{{\sc no} $>>$ {\sc some};} & {{\sc at most} $>$ {\sc between / exactly}} & {{\sc neg} $>$ {\sc few};} &  {{\sc no} $>$ {\sc fewer}}\\
{{\sc few} $>$ {\sc many};} & {{\sc few} $>$ {\sc between / exactly}} & {{\sc neg} $>$ {\sc fewer};} & {{\sc fewer} $>$ {\sc at most}}\\
{{\sc fewer} $>$ {\sc more};} & {{\sc fewer} $>$ {\sc between / exactly}} & {{\sc no} $>$ {\sc at most};} & {{\sc exactly} $>$ {\sc between}} \\
\hline
  \end{tabular}
  \caption{Graded monotonicity: summary of psycholinguistic experimental results \cite{geurts,sanford,chemla2011,yaron,milica}. The order in pairs represents that the first element is judged as a better NPI licenser than the second one or that it better supports DE inferences (or both). That is, `{\sc neg} $>>$ {\sc aff}' reads as `Sentences with sentential negation show much higher level of NPI acceptability or support DE inferences more than simple affirmative sentences.'. The `Logical monotonicity' side of the table groups together all relations expected under the logical view of monotonicity; `Subjective monotonicity' contains additional asymmetries found experimentally that do not follow from the simple logical view.}
  \label{predictions}
\end{table*}

The connection between monotonicity and NPI licensing is undeniable also beyond examples (1)-(3) (see \citealt{fauconnier,ladusaw} and much subsequent literature). Experimental evidence shows a bi-directional connection between inference judgments in a context and NPI acceptability in that context. \citet{chemla2011} found that the inferences a person considers valid in a given linguistic context predict how acceptable they would find an NPI in that same context. 
Conversely, \citet{milica} show that inferential judgments are modified by the presence of an NPI. So, the two phenomena show clear mutual influence. 

Importantly, both monotonicity and NPI acceptability in humans is not an all-or-nothing matter. Acceptance of logically valid inferences and rejection of invalid ones varies to some extent from person to person -- and from context to context \cite{geurts,sanford,chemla2011,yaron,milica}. 

\citet{chemla2011} report that logically DE sentences with {\it no} are perceived as DE by human participants only 72\% of the time. {\it At most} -- also logically a DE environment -- is only recognized as such 56\% of the time. Moreover, {\it less than} and {\it at most} -- truth-conditionally equivalent environments -- differ in DE inference endorsement by 11\%. The best predictor of NPI acceptability by humans was found to be not the logical entailment pattern but the subjective, or perceived, one \citep{chemla2011,milica}. 

There is no single overarching psycholinguistic study testing the whole landscape of contexts.
Combined knowledge from an array of studies \cite{geurts,sanford,chemla2011,yaron,milica} produces the picture summarized in Table \ref{predictions}. 

\section{Previous work}

NPIs have been a topic of an investigation in the context of LRMs, both as a subset of a more general test dataset \citep{marvin,hu2020systematic}, and as the main object of study \citep{jumelet,npi,illc,weber}. Here we focus on \citep{npi} as a representative case, as it shares with other previous studies its general set-up: assessment of LRMs against predictions of linguistic theory.

\citet{npi} focus on NPIs in BERT. Using a variety of testing techniques, both zero-shot and with fine-tuning, they conclude that BERT's ability to recognize NPI licensing environments and, therefore, to tell licit uses of NPIs from illicit ones varies a lot depending on the type of  context, scope configuration and the type of experimental setting. 

This might lead one to conclude that BERT's ability to recognize polarity of a sentence is not so great across the board. Indeed, reports from other tasks that involve polarity and/or monotonicity seem to support this. In particular, natural language inference has been reported to be hard for LRMs \citep{yanaka2019a,yanaka2019b,olmpics,potts}. Remarkably, \citet{potts} report that fine-tuning BERT on the SNLI dataset and then evaluating it on DE sentences (their NMoNLI dataset) results in 2.2\% accuracy -- that is, the model practically ignores the monotonicity profile of the sentence. But is alleged poor polarity detection to blame here?

Importantly for our study,  \citet{npi} judge BERT's recognition of NPI acceptability against logical monotonicity rather than subjective monotonicity as uncovered by psycholinguistic experiments. So, we believe that these results deserve a second look.

One of the measuring techniques in \citet{npi} is very close to one of the two techniques we will adopt in this paper. It is a version of Cloze Test adapted for MLM, where probabilities of candidates for the masked position are compared. We discuss the set-up in section 4.

Finally, the idea of targeted LRM evaluations modeled after {\bf psycholinguistic experiments} is being used in an increasing number of recent studies, albeit mainly in the domains of syntax and lexical semantics \citep{gulordava,linzen,marvin,wilcox,chowdhury,futrell,nair,abdou,ettinger}. 

We move on to describing our dataset, procedure and results.

\section{Method}
We perform two types of tests using the dataset that we produce for this purpose. One experiment is done with BERT, the other one with GPT-2. Both experiments are performed in a zero-shot setting -- using the pre-trained models without fine-tuning. The goal of these experiments is to test the contrasts between types of sentences described in Table \ref{predictions}. We will do this by comparing the relevant pairs of contexts along LRM-derived metrics that are meant to capture grammaticality / acceptability. 

First, we describe the dataset; then we explain the experiment procedure for BERT and GPT-2; finally, we report and discuss the results.

\subsection{The {\sc any} dataset\footnote{The data are available at \url{https://github.com/altsoph/Transformers-in-the-loop}}}

Our dataset consists of two parts: one with natural and one with synthetic data.

\subsubsection{Natural data}

We scraped the Gutenberg Project and a subset of English Wikipedia to obtain the list of sentences that contain {\it any}. Next, using a combination of heuristics\footnote{The script that can be used to reproduce the filtering procedure is available in the project repository, see fn. 2.}, we filtered the result with regular expressions to produce two sets of sentences (the second set underwent additional manual filtering):
\vspace*{-1ex}
\begin{itemize}
    \item 3844 sentences with sentential negation and a plural object with {\it any} to the right to the verb;
    \item 330 sentences with {\it nobody / no one} as subject and a plural object with {\it any} to the right.
\end{itemize}

\vspace*{-1ex}
\noindent The first set was modified to substitute the negated verb by its non-negated version, so we contrast 3844 sentences with negation and 3844 affirmative ones ({\sc neg} vs. {\sc aff}). In the second dataset, we substituted {\it nobody} for {\it somebody} and {\it no one} for {\it someone}, to check the {\sc some} vs. {\sc no} contrast.

\subsubsection{Synthetic data}

We used the following procedure. First, we automatically identified the set of verbs and nouns to build our items from. To do so, we started with {\tt bert-base-uncased}\footnote{ \url{https://huggingface.co/bert-base-uncased}} vocabulary. Taking its non-subword lexical tokens is an easy way to get a list of simple and common words. We ran this list through a SpaCy POS tagger\footnote{\url{https://github.com/explosion/spacy-models}}. Further, we lemmatized the result using {\tt pattern}\footnote{\url{https://pypi.org/project/Pattern/}} and dropped duplicates. Then, we filtered out modal verbs, singularia tantum nouns and some visible lemmatization mistakes. Finally, we filtered out non-transitive verbs to give the dataset a bit of a higher baseline of grammaticality.\footnote{Our procedure was equivalent to that in \url{github.com/Mirith/Verb-categorizer}} 

We kept top 100 nouns and top 100 verbs from the resulting lists -- these are the lexical entries we will deal with. Then, we generated sentences with these words, using the following pattern:

\begin{center}
A(n) {\tt noun}$_x$ {\tt verb.PST.SG} a(n) {\tt noun}$_y$\footnote{We use the singular indefinite object for this part of the procedure to avoid idiomatic verb phrases ({\it change hands}, {\it join forces}) at the top of the list.}
\end{center}

\noindent For this, we iterate over the 100 nouns in the subject and the object positions (excluding cases where the same noun appears in both positions) and over the 100 verbs. The procedure gave us 990k sentences like these:

\vspace*{-1ex}
\ex. \a. A girl crossed a road.
\b. A community hosted a game.
\b. A record put an air.

\vspace*{-1ex}
Some are more natural, make more sense and adhere to the verb's selectional restrictions better than the others. To control for this, we ran the sentences through {\tt GPT-2}\footnote{\url{https://huggingface.co/gpt2}} and assigned perplexity to all candidates. Then we took the bottom 20k of the sentences ($\approx$ the most `natural' ones) as the core of our synthetic dataset.

We tried to approximate the `naturalness' of examples by a combination of measures. We rely on  insights from different models (GPT-2, BERT,  corpus-based statistical insights into verb transitivity) on different stages of the dataset creation. Still, some sentences sound intuitively `weird'. We do not see this as a problem though -- we will not rely directly on the naturalness of individual examples, rather we will measure the effect of the NPI across the dataset (as is common practice when working with synthetic data -- see, for example, \citealt{potts,geiger2021causal}). The  amount of the examples will allow us to generalize across varying parts of the sentences to make sure that the results can be attributed to the parts we are interested in: items responsible for the polarity of the sentence. The quantity of test items is crucial for reproducing psycholinguistic experiments on LRMs -- while in the former one sentence gives rise to a number of observations when different human participants make a judgment, in the latter one test sentence gives one observation only.

With this in mind, we use the 20k sentences produced by the previous steps to build the parts of our synthetic dataset. Each of the sentences has a pluralized (not singular anymore!)  object in combination with {\it any}: {\tt any roads}. The subject type varies in different datasets comprising our synthetic data. Here is what we end up with:

\vspace*{-1ex}
\begin{itemize}
    \item 12 datasets 20k sentences each: \\{\sc aff} \Next[a]; {\sc neg} \Next[b]; {\sc some} \Next[c]; {\sc no}; {\sc many}; {\sc few}; {\sc more than 5}; {\sc fewer than 5}; {\sc at least 5}; {\sc at most 5}; {\sc exactly 5}; {\sc between 5 and 10};
    \item 2 datasets 8230 sentences each: \\{\sc somebody / someone / something} \Next[d]; {\sc nobody / no one / nothing} (replacing the whole subject, duplicates deleted)
\end{itemize}

\vspace*{-2ex}
\ex. \a. A girl crossed any roads.
\b. A girl didn't cross any roads.
\b. Some girls crossed any roads.
\b. Somebody crossed any roads.

\vspace*{-1ex}
Overall, sentences in all parts of our dataset vary in the type of context it instantiates (simple affirmative, negation, different quantifiers) -- but all sentences contain {\it any} in the object position in combination with a plural noun.

The next two subsections explain the metrics derived from the two model we study, stemming from the differences in their architecture and training objectives.

\begin{figure*}[h]
 \subfloat[\centering BERT-prob comparison across conditions]{{\includegraphics[width=.499\textwidth]{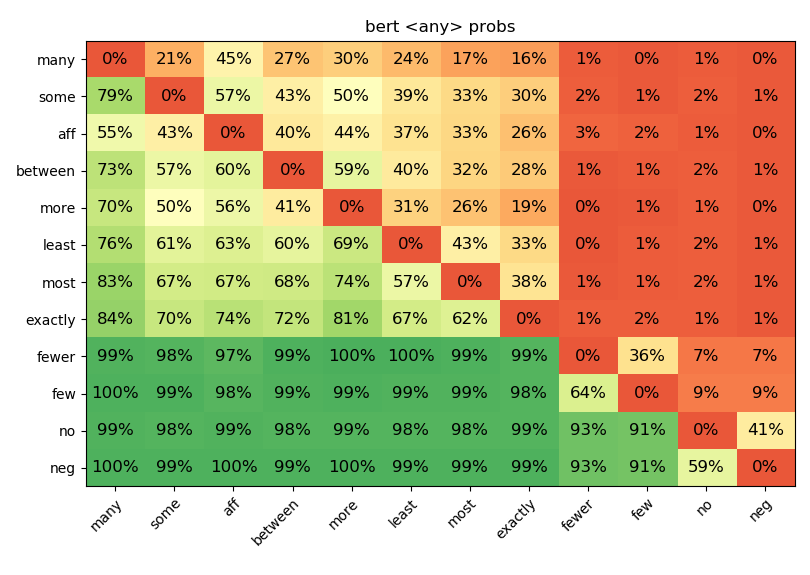} }}
\subfloat[\centering GPT-PPL-diff comparison across conditions]{{\includegraphics[width=.463\textwidth]{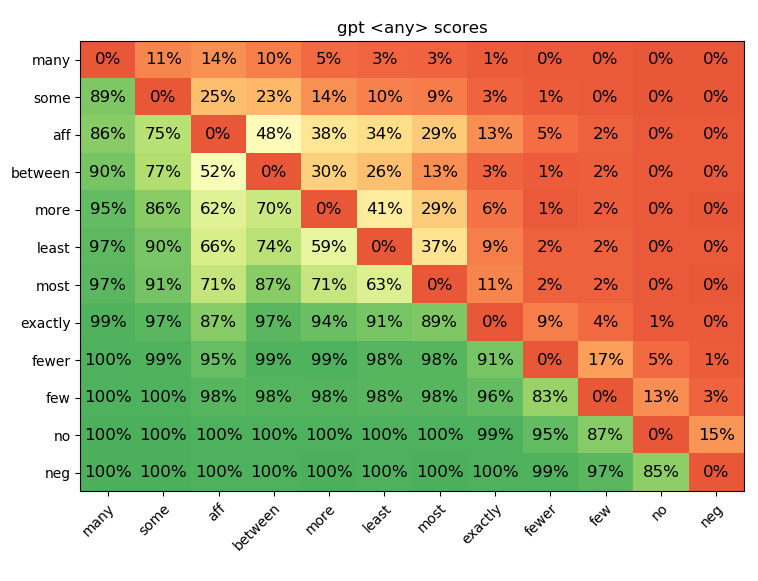} }}
\caption{LRM experiment results}
\label{lrm_results}
\end{figure*}

\subsection{BERT: Cloze Test}
The Cloze Test on BERT is very similar to that described in \cite{npi}. In each of the sentences in the dataset, we mask {\it any} and ask BERT for predictions for the masked position:

\begin{center}
{\small{\tt [CLS]\hspace{.2em plus .3em}Few\hspace{.5em plus .6em}girls\hspace{.5em plus .6em}crossed\hspace{.3em plus .4em}[MASK]\hspace{.2em plus .3em}roads\hspace{.2em plus .3em}.\hspace{.2em plus .3em}[SEP]}}
\end{center}

\noindent We extract the probability that BERT assigns to {\it any} in the masked position, as well as the rank of {\it any} in BERT vocabulary sorted by the probability in the masked position. 

Further, we compare these values between conditions (= different types of contexts). The comparison between a pair of conditions will be expressed as the percentage of sentences in our dataset where {\it any} got a higher probability in the first condition compared to the probability of {\it any} in the corresponding sentence in the second condition. The same for the rank of {\it any} instead of probability. For example, $\langle \textsc{aff: neg} \rangle: 0.12\% $ reads as: in 0.12\% of the dataset, {\it any} got a higher probability (or a higher rank) in an affirmative sentence compared to the corresponding sentence with negation. Intuitively: that most of the time, a sentence with negation makes a better environment for {\it any} than the minimally different affirmative sentence.

\subsection{GPT-2: Perplexity difference}

In this test, for each sentence in the dataset, we calculate perplexity of this sentence \Next[a] according to the GPT-2 model -- and perplexity of that same sentence with {\it any} deleted \Next[b]:

\vspace*{-1ex}
\ex. \a. Few girls crossed any roads.
\b. Few girls crossed roads.

\vspace*{-1ex}
We take the difference between these perplexity values normalized by the number of tokens as our measure of how much the presence of {\it any} affects the `naturalness' of each particular sentence.

As before, we compare these values for different conditions. For example, $\langle \textsc{aff: neg} \rangle: 0.25\% $ reads as: in 0.25\% of sentences, the presence of {\it any} leads to a smaller increase in perplexity for the affirmative sentence, compared to the analogous negative sentence. That is, most of the time the presence of {\it any} worsens affirmative sentences a lot, while the corresponding negative one -- less so. 

This is the closest possible LM analogue of the acceptability judgment experiments like \citep{sub2019}, which measure the differences between acceptability scores with and without {\it any} for different types of contexts.

\section{Results of model evaluation}

We will discuss results from BERT and GPT-2 together, because they mostly agree.

One general result that allows us to limit our attention to one of the two BERT metrics is that BERT rank and BERT probability produce the same order on all condition pairs of interest except for one ($\langle \textsc{at most, at least} \rangle$) and we will only discuss BERT probabilities in this section.

The 20k synthetic data results are summarized in Fig. \ref{lrm_results}. The conditions in the 20k results are sorted for readability. 8k synthetic data results: $\langle \textsc{no-},\textsc{some-} \rangle$: 99.76\% (BERT-prob); 99.56\% (GPT-PPL-diff).

In short, {\bf all predictions based on psycholinguistic evidence discussed in section 2 (Table \ref{predictions}) are confirmed by our LRM data.}

As a sanity check, we compare these results with the results of the same procedure on our natural dataset, and they are very similar: $\langle \textsc{neg, aff} \rangle$: 97.21\% (BERT-prob), 97.17\% (GPT-PPL-diff); $\langle \textsc{no-, some-} \rangle$: 98.29\% (BERT-prob), 96.98\% (GPT-PPL-diff). 

The take home message from these results is that {\bf LRMs can tell between negative and positive polarity, as well as between different types of contexts by their monotonicity, as measured by NPI acceptability}. Moreover, what is encoded is a subjective version of the relevant property, similar to what is reflected in graded non-categorical judgments seen in psycholinguistic experiments.

Establishing this, first of all, helps us make more sense of the metrics derived from such models and helps draw a more accurate line between noise and meaningful output. Second, it encourages a closer tie between experiments with humans and with LRMs: LRMs encode a snapshot of numerous subjective linguistic intuitions, and maybe we can use LRMs to get indirect access to speakers' shared intuitions as a source of new theoretically relevant linguistic generalisations. 
The next section is a pilot attempt in this direction. We establish a new generalization looking at LRM data -- and then confirm it in a  psycholinguistic experiment. 

\section{Next step: Cardinality dependency}
For the conditions which involve numerals we left one parameter unexplored so far, namely, the numeral itself. 
 In this section, we look at the dependency between NPI acceptability and the numeral.

There is no experimental data on this. Theoretical literature tentatively suggests that the higher the numeral, the less acceptable an NPI in its scope \cite{crnic}:

\vspace*{-1ex}
\ex. Exactly two of the boxes contain anything

\vspace*{-4ex}
\ex. $^{??}$Exactly 98 of the boxes contain anything

\vspace*{-1ex}
However, the judgments are subtle and theoretical discussion still waits for an empirical basis. 
Let us look at our conditions with numerals (apart from {\sc between} -- we set it aside as too complicated). For each of the conditions, we keep everything constant apart from the numeral and check the effect the numeral has on NPI acceptability.

\subsection{As seen in LRMs}

We looked at numerals with these numeric values: $[2\!-\!20,30,40,50,60,70,80,90]$. As before, we made pair-wise comparisons between sentences in our synthetic dataset that differ only in the numeral it contains. The measures are the same as before.

\begin{figure}[h!]
   \centering
   \includegraphics[width=.45\textwidth]{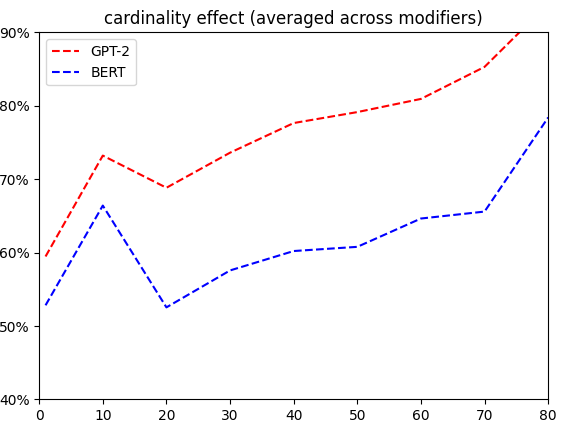} 
   \caption{The effect of numeral on {\tt any}.}
   \label{num_dep}
\end{figure}

Both models show an upward trend: the higher the numeral, the worse the context becomes for {\it any}. This tendency is shown on Fig. \ref{num_dep}.


The lines show comparison between sentence pairs in which the second one has a numeral higher than the one in the first sentence by $n$, where $n$ is plotted on the $x$ axis (so, $10$ on the $x$ axis comprises all pairs that differ by $10$ -- $\langle 2,12 \rangle$,$\langle 3,13 \rangle$...). On $y$, we show the percentage of pairs in which the first sentence showed higher probability of {\it any} than the second one.

\noindent The effect of the numeral on the NPI acceptability can be sometimes quite strong: to the point of flipping the `better NPI licenser' relation in a pair of contexts. For example, this is the case for {\sc at least} and {\sc more than} in BERT. They have the same logical monotonicity profile (both UE). 
However, we can find a pair of numerals such that flipping them orders the resulting contexts differently:

\begin{center}
    {\sc at least 2 $>$ more than 70}: 94\% \\
{\sc more than 2 $>$ at least 70}: 68\%
\end{center}

\noindent Let us check the effect of numeral on humans, as well as a licensing flip due to the numeral.

\subsection{In humans}

For the ease of comparison between our LRM experiment data in the previous section and the experiment on human participants, we formulate the latter as a {\bf forced-choice task}.

The participants saw pairs of sentences and were instructed to pick the one that is more grammatical. The study has a {\bf 2x2 design} with these factors:
\vspace*{-1ex}
\begin{itemize}
    \item {\sc numeral}: {\it five} vs. {\it seventy}
    \vspace*{-2ex}
    \item {\sc quantifier}: {\it at least} vs. {\it more than}
\end{itemize}

\vspace*{-1ex}
\noindent This gives six forced-choice test conditions: 
\begin{center}
    {\it at least five} vs. {\it at least seventy} \\
    {\it at least five} vs. {\it more than five} \\ 
    {\it at least five} vs. {\it more than seventy} \\
    {\it at least seventy} vs. {\it more than five}\\ 
    {\it at least seventy} vs. {\it more than seventy}\\
    {\it more than five} vs. {\it more than seventy}
\end{center}

\noindent These prefixes were used to generate pairs of sentences using patterns from the 20k synthetic dataset. 
%
%
We randomly selected 50 out of the 20k patterns, which results in 2500 pattern pairs. With 6 test conditions, this amounts to {\bf 15k unique test items}. 

We used Toloka to recruit self-reported native speakers of English for this experiment.\footnote{\url{https://toloka.ai/ready-to-go/}} They were allowed to complete the full task after they passed a test with 10 control items with 7 or more correctly identified grammatical sentences. 

In the main part of the task, each participant saw {\bf 38 pairs of sentences}: 22 were filler/control items and 16 test items. All participants saw the same filler/control items (random order), test items were taken from the pool of 15k test items in random order and evaluated with no overlap. 

\begin{figure}[h!]
   \centering
   \includegraphics[width=.5\textwidth]{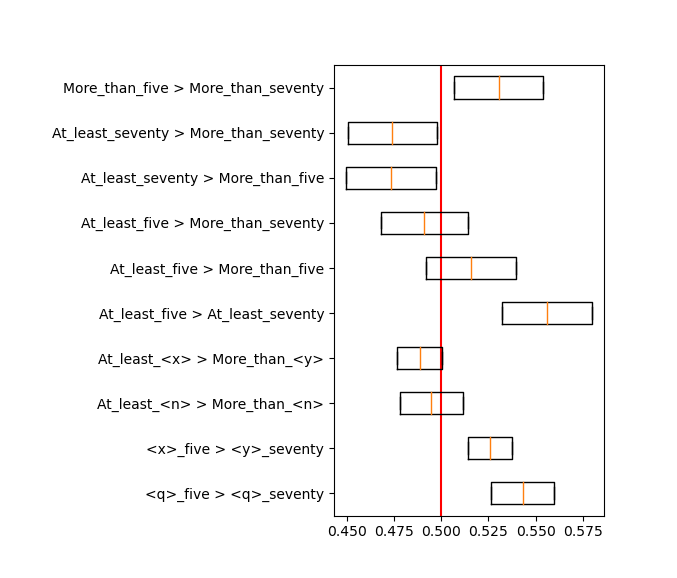} 
   \caption{Human judgments of {\it any}-acceptability}
   \label{humans}
\end{figure}

In total, 968 participants were recruited. We filtered out the data from those who gave wrong answers to more than 30\% of the filter/control items in the main part of the task. We were left with 656 participants (= 10496 test items; more than a 2/3 of our pool of test items).  Fig. \ref{humans} shows the {\bf results} of the experiment. We used the binomial test to analyze the data. The boxes in the plot show the 95\% confidence interval.

\noindent{\bf Result \#1}: The effect of the numeral is confirmed both within and across the two types of contexts (lines 1, 6, 9 and 10 in Fig. \ref{humans}). {\bf Result \#2:} {\sc at least} and {\sc more than} are not ordered with respect to each other (lines 7 and 8). It is possible to find a particular numeral where the difference reaches significance (line 2), but overall there is no clear order. {\bf Result \#3:} Our data do not show a statistically significant flip between contexts with different numeral values. Even though one side of the flip is there (line 3), the flip of this pair did not reach significance (line 5).  

\noindent{\bf Conclusion:} The results are generally in line with the trend observed in section 6: the higher the numeral, the worse the context gets for an NPI. This is the first experimental confirmation of this effect, to the best of our knowledge. It is noteworthy that we first found it via LRM -- and then confirmed it with human participants. 

A more specific result of this effect -- what we call a `flip' -- is seen in our data as a tendency, but the effect did not reach significance. It could be an LRM artifact -- or the lack of it could be an artifact of our experiment. A different choice of numerals or a higher number of participants could sharpen these results. We leave this for future work.

\section{Discussion and outlook}

Our experiments provide solid support for an approach under which LRM performance is compared directly to psycholinguistic data rather than to predictions of a linguistic theory. This opens up prospects for research that will result in a more empirically grounded picture of where the limits of LRM abilities lie. 

Our results tell us something new about LRMs but also suggest that LRMs can be included in the experimental loop of theoretical semantics alongside with traditional experiments. To pilot this idea, we conducted an experiment on the effect of the numeral on NPI acceptability. We confirmed our LRM findings in a parallel psycholinguistic study. 

In this paper, we only explore the connection between behavioral experiments and LRM-derived metrics. What about online measures in psycholinguistic studies? Can we find a usable analogue to, for example, eye-tracking or reaction times in self-paced reading studies -- that is, studies that tell us which parts of input are important in processing? One obvious LRM-based candidate is attention. 

We took a preliminary look at BERT attention distribution in sentences with {\it any} in an attempt to identify the attention head that contributes most to monotonicity-via-NPIs (see \citealt{voita} for a discussion of attention head specialization). To factor out linear position, we focused on the natural part of our dataset. We took the sentences that contained both a quantifier with a clear monotonicity profile ({\it somebody}, {\it nobody}, {\it someone} etc.) and {\it any}; calculated attention from {\it any} to the quantifier for every layer and every attention head and averaged it across sentences. Then we sorted the results and went through the top of the resulting list. 

We found that the attention head (6,2) of {\tt bert-base-uncased} model -- 6th layer, attention head 2 -- seems to specialize in precisely what we are looking for. Saliency maps below show that in a variety of contexts beyond the ones we checked for the purposes of this paper, monotonicity-affecting items are highlighted -- buttressing the hypothesis that monotonicity is important for NPI licensing ({\it without}, {\it do}-support in a question, {\it if}, lexical negation):

\vspace{1ex}
\begin{flushleft}{\tiny
\texttt{
\colorbox[HTML]{FFFCFC}{[CLS]}\hspace{.01em plus .001em}\colorbox[HTML]{FFFBFB}{it}\hspace{.01em plus .001em}\colorbox[HTML]{FFFAFA}{felt}\hspace{.01em plus .001em}\colorbox[HTML]{FFE8E8}{odd}\hspace{.01em plus .001em}\colorbox[HTML]{FFBDBD}{without}\hspace{.01em plus .001em}\colorbox[HTML]{FFEBEB}{any}\hspace{.01em plus .001em}\colorbox[HTML]{FFFBFB}{wards}\hspace{.01em plus .001em}\colorbox[HTML]{FFFFFF}{on}\hspace{.01em plus .001em}\colorbox[HTML]{FFFFFF}{it}\hspace{.01em plus .001em}\colorbox[HTML]{FFFBFB}{.}\hspace{.01em plus .001em}\colorbox[HTML]{FF8787}{[SEP]}
}}
\end{flushleft}
\begin{flushleft}{\tiny
\texttt{
\colorbox[HTML]{FFFAFA}{[CLS]}\hspace{.01em plus .001em}\colorbox[HTML]{FF7B7B}{do}\hspace{.01em plus .001em}\colorbox[HTML]{FFF9F9}{you}\hspace{.01em plus .001em}\colorbox[HTML]{FFFBFB}{have}\hspace{.01em plus .001em}\colorbox[HTML]{FFFAFA}{any}\hspace{.01em plus .001em}\colorbox[HTML]{FFFDFD}{brothers}\hspace{.01em plus .001em}\colorbox[HTML]{FFFFFF}{or}\hspace{.01em plus .001em}\colorbox[HTML]{FFFEFE}{sisters}\hspace{.01em plus .001em}\colorbox[HTML]{FFF7F7}{?}\hspace{.01em plus .001em}\colorbox[HTML]{FFA8A8}{[SEP]}
}}
\end{flushleft}
\begin{flushleft}{\tiny
\texttt{
\colorbox[HTML]{FFFEFE}{[CLS]}\hspace{.01em plus .001em}\colorbox[HTML]{FF3535}{if}\hspace{.01em plus .001em}\colorbox[HTML]{FFFEFE}{there}\hspace{.01em plus .001em}\colorbox[HTML]{FFFDFD}{'}\hspace{.01em plus .001em}\colorbox[HTML]{FFF6F6}{d}\hspace{.01em plus .001em}\colorbox[HTML]{FFFDFD}{been}\hspace{.01em plus .001em}\colorbox[HTML]{FFFDFD}{any}\hspace{.01em plus .001em}\colorbox[HTML]{FFFFFF}{babies}\hspace{.01em plus .001em}\colorbox[HTML]{FFFFFF}{present}\hspace{.01em plus .001em}\colorbox[HTML]{FFFFFF}{,}\hspace{.01em plus .001em}\colorbox[HTML]{FFFFFF}{he}\hspace{.01em plus .001em}\colorbox[HTML]{FFFFFF}{'}\hspace{.01em plus .001em}\colorbox[HTML]{FFFDFD}{d}\hspace{.01em plus .001em}\colorbox[HTML]{FFFFFF}{have}\hspace{.01em plus .001em}\colorbox[HTML]{FFFFFF}{been}\hspace{.01em plus .001em}\colorbox[HTML]{FFFFFF}{un}\hspace{.01em plus .001em}\colorbox[HTML]{FFFFFF}{\#\#sto}\hspace{.01em plus .001em}\colorbox[HTML]{FFFFFF}{\#\#ppa}\hspace{.01em plus .001em}\colorbox[HTML]{FFFFFF}{\#\#ble}\hspace{.01em plus .001em}\colorbox[HTML]{FFFEFE}{.}\hspace{.01em plus .001em}\colorbox[HTML]{FFE6E6}{[SEP]}
}}
\end{flushleft}
\begin{flushleft}{\tiny
\texttt{
\colorbox[HTML]{FFF9F9}{[CLS]}\hspace{.01em plus .001em}\colorbox[HTML]{FFE8E8}{we}\hspace{.01em plus .001em}\colorbox[HTML]{FFEBEB}{are}\hspace{.01em plus .001em}\colorbox[HTML]{FFCDCD}{unable}\hspace{.01em plus .001em}\colorbox[HTML]{FFF3F3}{to}\hspace{.01em plus .001em}\colorbox[HTML]{FFEFEF}{identify}\hspace{.01em plus .001em}\colorbox[HTML]{FFEDED}{any}\hspace{.01em plus .001em}\colorbox[HTML]{FFF4F4}{others}\hspace{.01em plus .001em}\colorbox[HTML]{FFFCFC}{who}\hspace{.01em plus .001em}\colorbox[HTML]{FFFFFF}{knew}\hspace{.01em plus .001em}\colorbox[HTML]{FFFFFF}{of}\hspace{.01em plus .001em}\colorbox[HTML]{FFFFFF}{the}\hspace{.01em plus .001em}\colorbox[HTML]{FFFFFF}{scheme}\hspace{.01em plus .001em}\colorbox[HTML]{FFFFFF}{at}\hspace{.01em plus .001em}\colorbox[HTML]{FFFFFF}{the}\hspace{.01em plus .001em}\colorbox[HTML]{FFFFFF}{time}\hspace{.01em plus .001em}\colorbox[HTML]{FFFFFF}{it}\hspace{.01em plus .001em}\colorbox[HTML]{FFFFFF}{was}\hspace{.01em plus .001em}\colorbox[HTML]{FFFFFF}{being}\hspace{.01em plus .001em}\colorbox[HTML]{FFFFFF}{considered}\hspace{.01em plus .001em}\colorbox[HTML]{FFFEFE}{.}\hspace{.01em plus .001em}\colorbox[HTML]{FFA7A7}{[SEP]}
}}
\end{flushleft}

\vspace{1ex}

\noindent Additionally, this attention head reflects the role of the numeral in NPI licensing that we established in section 6: in all contexts with numerals that we looked at, a lot of attention goes from {\it any} to both the quantifier (say, {\it exactly}) and the numeral that comes with it. Moreover, the higher the numeral, the more attention goes to it, compared to the amount of attention that goes to the quantifier:

\vspace{.5ex}
\begin{flushleft}{\tiny
\texttt{
\colorbox[HTML]{FFF2F2}{[CLS]}\hspace{.01em plus .001em}\colorbox[HTML]{FFA5A5}{exactly}\hspace{.01em plus .001em}\colorbox[HTML]{FFDADA}{two}\hspace{.01em plus .001em}\colorbox[HTML]{FFF4F4}{games}\hspace{.01em plus .001em}\colorbox[HTML]{FFF6F6}{told}\hspace{.01em plus .001em}\colorbox[HTML]{FFF6F6}{any}\hspace{.01em plus .001em}\colorbox[HTML]{FFFEFE}{stories}\hspace{.01em plus .001em}\colorbox[HTML]{FFF1F1}{.}\hspace{.01em plus .001em}\colorbox[HTML]{FFBEBE}{[SEP]}
}}
\end{flushleft}

\begin{flushleft}{\tiny
\texttt{
\hspace{.01em plus .001em}\colorbox[HTML]{FFF7F7}{[CLS]}\hspace{.01em plus .001em}\colorbox[HTML]{FFE3E3}{exactly}\hspace{.01em plus .001em}\colorbox[HTML]{FF9C9C}{ninety}\hspace{.02em plus .001em}\colorbox[HTML]{FFF8F8}{games}\hspace{.01em plus .001em}\colorbox[HTML]{FFF9F9}{told}\hspace{.01em plus .001em}\colorbox[HTML]{FFF9F9}{any}\hspace{.01em plus .001em}\colorbox[HTML]{FFFEFE}{stories}\hspace{.01em plus .001em}\colorbox[HTML]{FFF4F4}{.}\hspace{.01em plus .001em}\colorbox[HTML]{FFABAB}{[SEP]}
}}
\end{flushleft}
\vspace{.5ex}

\noindent More work is needed to verify and interpret these patterns systematically and compare them to other attribution measures and to online metrics in psycholinguistic studies. 

\section*{Acknowledgements}

We thank the anonymous ARR reviewers; the  audience and organizers of the CNRS Seminar on the Interactions between Formal and Computational Linguistics; Toloka team for the help with the human assessment study. We also thank Alexandre Cremers, Ekaterina Garmash, Borislav Kozlovskii, Rick Nouwen, and Denis Paperno for the discussions of our ideas and earlier versions of the paper.

\bibliographystyle{acl_natbib}
\bibliography{references}

\begin{thebibliography}{36}
\expandafter\ifx\csname natexlab\endcsname\relax\def\natexlab#1{#1}\fi

\bibitem[{Abdou et~al.(2020)Abdou, Ravishankar, Barrett, Belinkov, Elliott, and
  S{\o}gaard}]{abdou}
Mostafa Abdou, Vinit Ravishankar, Maria Barrett, Yonatan Belinkov, Desmond
  Elliott, and Anders S{\o}gaard. 2020.
\newblock \href {https://doi.org/10.18653/v1/2020.acl-main.679} {The
  sensitivity of language models and humans to {W}inograd schema
  perturbations}.
\newblock In \emph{Proceedings of the 58th Annual Meeting of the Association
  for Computational Linguistics}, pages 7590--7604.

\bibitem[{Alexandropoulou et~al.(2020)Alexandropoulou, Bylinina, and
  Nouwen}]{sub2019}
Stavroula Alexandropoulou, Lisa Bylinina, and Rick Nouwen. 2020.
\newblock Is there `any' licensing in non-{DE} contexts? {A}n experimental
  study.
\newblock In \emph{Proceedings of Sinn und Bedeutung}, volume~24, pages 35--47.

\bibitem[{Barker(2018)}]{barker}
Chris Barker. 2018.
\newblock Negative polarity as scope marking.
\newblock \emph{Linguistics and philosophy}, 41(5):483--510.

\bibitem[{Baroni(2021)}]{baroni2021}
Marco Baroni. 2021.
\newblock On the proper role of linguistically-oriented deep net analysis in
  linguistic theorizing.
\newblock \emph{arXiv preprint arXiv:2106.08694}.

\bibitem[{Chemla et~al.(2011)Chemla, Homer, and Rothschild}]{chemla2011}
Emmanuel Chemla, Vincent Homer, and Daniel Rothschild. 2011.
\newblock Modularity and intuitions in formal semantics: The case of polarity
  items.
\newblock \emph{Linguistics and Philosophy}, 34(6):537--570.

\bibitem[{Chowdhury and Zamparelli(2018)}]{chowdhury}
Shammur~Absar Chowdhury and Roberto Zamparelli. 2018.
\newblock {RNN} simulations of grammaticality judgments on long-distance
  dependencies.
\newblock In \emph{Proceedings of the 27th international conference on
  computational linguistics}, pages 133--144.

\bibitem[{Crni{\v{c}}(2014)}]{crnic}
Luka Crni{\v{c}}. 2014.
\newblock Non-monotonicity in {NPI} licensing.
\newblock \emph{Natural Language Semantics}, 22(2):169--217.

\bibitem[{Denić et~al.(2020)Denić, Homer, Rothschild, and Chemla}]{milica}
Milica Denić, Vincent Homer, Daniel Rothschild, and Emmanuel Chemla. 2020.
\newblock The influence of polarity items on inferential judgments.
\newblock Submitted.

\bibitem[{Devlin et~al.(2019)Devlin, Chang, Lee, and Toutanova}]{bert}
Jacob Devlin, Ming-Wei Chang, Kenton Lee, and Kristina Toutanova. 2019.
\newblock {BERT}: Pre-training of deep bidirectional transformers for language
  understanding.
\newblock In \emph{NAACL-HLT}.

\bibitem[{Ettinger(2020)}]{ettinger}
Allyson Ettinger. 2020.
\newblock What {BERT} is not: Lessons from a new suite of psycholinguistic
  diagnostics for language models.
\newblock volume~8, pages 34--48. MIT Press.

\bibitem[{Fauconnier(1975)}]{fauconnier}
Gilles Fauconnier. 1975.
\newblock Polarity and the scale principle.
\newblock In \emph{Proceedings of {C}hicago {L}inguistc {S}ociety 11}, pages
  188--99.

\bibitem[{Futrell et~al.(2019)Futrell, Wilcox, Morita, Qian, Ballesteros, and
  Levy}]{futrell}
Richard Futrell, Ethan Wilcox, Takashi Morita, Peng Qian, Miguel Ballesteros,
  and Roger Levy. 2019.
\newblock \href {https://doi.org/10.18653/v1/N19-1004} {Neural language models
  as psycholinguistic subjects: Representations of syntactic state}.
\newblock In \emph{Proceedings of the 2019 Conference of the North {A}merican
  Chapter of the Association for Computational Linguistics: Human Language
  Technologies, Volume 1 (Long and Short Papers)}, pages 32--42.

\bibitem[{Geiger et~al.(2021)Geiger, Lu, Icard, and Potts}]{geiger2021causal}
Atticus Geiger, Hanson Lu, Thomas Icard, and Christopher Potts. 2021.
\newblock Causal abstractions of neural networks.
\newblock \emph{arXiv preprint arXiv:2106.02997}.

\bibitem[{Geiger et~al.(2020)Geiger, Richardson, and Potts}]{potts}
Atticus Geiger, Kyle Richardson, and Christopher Potts. 2020.
\newblock Neural natural language inference models partially embed theories of
  lexical entailment and negation.
\newblock In \emph{Proceedings of the Third BlackboxNLP Workshop on Analyzing
  and Interpreting Neural Networks for NLP}, pages 163--173.

\bibitem[{Geurts(2003)}]{geurts}
Bart Geurts. 2003.
\newblock Reasoning with quantifiers.
\newblock \emph{Cognition}, 86(3):223--251.

\bibitem[{Giannakidou(1998)}]{anastasia}
Anastasia Giannakidou. 1998.
\newblock \emph{Polarity sensitivity as (non) veridical dependency}, volume~23.
\newblock John Benjamins Publishing.

\bibitem[{Goldberg(2019)}]{goldberg}
Yoav Goldberg. 2019.
\newblock Assessing {BERT}’s syntactic abilities.
\newblock \emph{arXiv preprint arXiv:1901.05287}.

\bibitem[{Gulordava et~al.(2018)Gulordava, Bojanowski, Grave, Linzen, and
  Baroni}]{gulordava}
Kristina Gulordava, Piotr Bojanowski, Edouard Grave, Tal Linzen, and Marco
  Baroni. 2018.
\newblock \href {https://doi.org/10.18653/v1/N18-1108} {Colorless green
  recurrent networks dream hierarchically}.
\newblock In \emph{Proceedings of the 2018 Conference of the North {A}merican
  Chapter of the Association for Computational Linguistics: Human Language
  Technologies, Volume 1 (Long Papers)}, pages 1195--1205.

\bibitem[{Hu et~al.(2020)Hu, Gauthier, Qian, Wilcox, and
  Levy}]{hu2020systematic}
Jennifer Hu, Jon Gauthier, Peng Qian, Ethan Wilcox, and Roger Levy. 2020.
\newblock A systematic assessment of syntactic generalization in neural
  language models.
\newblock In \emph{Proceedings of the 58th Annual Meeting of the Association
  for Computational Linguistics}, pages 1725--1744.

\bibitem[{Jumelet et~al.(2021)Jumelet, Denic, Szymanik, Hupkes, and
  Steinert{-}Threlkeld}]{illc}
Jaap Jumelet, Milica Denic, Jakub Szymanik, Dieuwke Hupkes, and Shane
  Steinert{-}Threlkeld. 2021.
\newblock \href {http://arxiv.org/abs/2105.13818} {Language models use
  monotonicity to assess {NPI} licensing}.
\newblock \emph{CoRR}, abs/2105.13818.

\bibitem[{Jumelet and Hupkes(2018)}]{jumelet}
Jaap Jumelet and Dieuwke Hupkes. 2018.
\newblock Do language models understand anything? on the ability of lstms to
  understand negative polarity items.
\newblock In \emph{BlackboxNLP@ EMNLP}.

\bibitem[{Ladusaw(1979)}]{ladusaw}
William~A Ladusaw. 1979.
\newblock \emph{Polarity sensitivity as inherent scope relations}.
\newblock Ph.D. thesis, Austin, TX: University of Texas at Austin.

\bibitem[{Linzen and Baroni(2021)}]{linzen2021}
Tal Linzen and Marco Baroni. 2021.
\newblock Syntactic structure from deep learning.
\newblock \emph{Annual Review of Linguistics}, 7:195--212.

\bibitem[{Linzen et~al.(2016)Linzen, Dupoux, and Goldberg}]{linzen}
Tal Linzen, Emmanuel Dupoux, and Yoav Goldberg. 2016.
\newblock Assessing the ability of {LSTM}s to learn syntax-sensitive
  dependencies.
\newblock In \emph{Transactions of the Association for Computational
  Linguistics}, volume~4, pages 521--535. MIT Press.

\bibitem[{Marvin and Linzen(2018)}]{marvin}
Rebecca Marvin and Tal Linzen. 2018.
\newblock \href {https://doi.org/10.18653/v1/D18-1151} {Targeted syntactic
  evaluation of language models}.
\newblock In \emph{Proceedings of the 2018 Conference on Empirical Methods in
  Natural Language Processing}, pages 1192--1202.

\bibitem[{McNabb et~al.(2016)McNabb, Alexandropoulou, Blok, Bimpikou, and
  Nouwen}]{yaron}
Yaron McNabb, Stavroula Alexandropoulou, Dominique Blok, Sofia Bimpikou, and
  Rick Nouwen. 2016.
\newblock The likelihood of upper-bound construals among numeral modifiers.
\newblock In \emph{Proceedings of Sinn und Bedeutung}, volume~20, pages
  497--514.

\bibitem[{Nair et~al.(2020)Nair, Srinivasan, and Meylan}]{nair}
Sathvik Nair, Mahesh Srinivasan, and Stephan Meylan. 2020.
\newblock \href {https://www.aclweb.org/anthology/2020.cogalex-1.16}
  {Contextualized word embeddings encode aspects of human-like word sense
  knowledge}.
\newblock In \emph{Proceedings of the Workshop on the Cognitive Aspects of the
  Lexicon}, pages 129--141.

\bibitem[{Radford et~al.(2019)Radford, Wu, Child, Luan, Amodei, and
  Sutskever}]{gpt}
Alec Radford, Jeff Wu, Rewon Child, David Luan, Dario Amodei, and Ilya
  Sutskever. 2019.
\newblock Language models are unsupervised multitask learners.

\bibitem[{Sanford et~al.(2007)Sanford, Dawydiak, and Moxey}]{sanford}
Anthony~J Sanford, Eugene~J Dawydiak, and Linda~M Moxey. 2007.
\newblock A unified account of quantifer perspective effects in discourse.
\newblock \emph{Discourse Processes}, 44(1):1--32.

\bibitem[{Talmor et~al.(2020)Talmor, Elazar, Goldberg, and Berant}]{olmpics}
Alon Talmor, Yanai Elazar, Yoav Goldberg, and Jonathan Berant. 2020.
\newblock o{LM}pics-on what language model pre-training captures.
\newblock In \emph{Transactions of the Association for Computational
  Linguistics}, volume~8, pages 743--758. MIT Press.

\bibitem[{Voita et~al.(2019)Voita, Talbot, Moiseev, Sennrich, and
  Titov}]{voita}
Elena Voita, David Talbot, Fedor Moiseev, Rico Sennrich, and Ivan Titov. 2019.
\newblock \href {https://doi.org/10.18653/v1/P19-1580} {Analyzing multi-head
  self-attention: Specialized heads do the heavy lifting, the rest can be
  pruned}.
\newblock In \emph{Proceedings of the 57th Annual Meeting of the Association
  for Computational Linguistics}, pages 5797--5808, Florence, Italy.
  Association for Computational Linguistics.

\bibitem[{Warstadt et~al.(2019)Warstadt, Cao, Grosu, Peng, Blix, Nie, Alsop,
  Bordia, Liu, Parrish et~al.}]{npi}
Alex Warstadt, Yu~Cao, Ioana Grosu, Wei Peng, Hagen Blix, Yining Nie, Anna
  Alsop, Shikha Bordia, Haokun Liu, Alicia Parrish, et~al. 2019.
\newblock \href {https://doi.org/10.18653/v1/D19-1286} {Investigating bert's
  knowledge of language: Five analysis methods with npis}.
\newblock In \emph{Proceedings of the 2019 Conference on Empirical Methods in
  Natural Language Processing and the 9th International Joint Conference on
  Natural Language Processing (EMNLP-IJCNLP)}, pages 2877--2887.

\bibitem[{Weber et~al.(2021)Weber, Jumelet, Bruni, and Hupkes}]{weber}
Lucas Weber, Jaap Jumelet, Elia Bruni, and Dieuwke Hupkes. 2021.
\newblock \href {https://doi.org/10.18653/v1/2021.eacl-main.176} {Language
  modelling as a multi-task problem}.
\newblock In \emph{Proceedings of the 16th Conference of the European Chapter
  of the Association for Computational Linguistics: Main Volume}, pages
  2049--2060, Online. Association for Computational Linguistics.

\bibitem[{Wilcox et~al.(2018)Wilcox, Levy, Morita, and Futrell}]{wilcox}
Ethan Wilcox, Roger Levy, Takashi Morita, and Richard Futrell. 2018.
\newblock \href {https://doi.org/10.18653/v1/W18-5423} {What do {RNN} language
  models learn about filler-gap dependencies?}
\newblock In \emph{Proceedings of the 2018 {EMNLP} Workshop {B}lackbox{NLP}:
  Analyzing and Interpreting Neural Networks for {NLP}}, pages 211--221.

\bibitem[{Yanaka et~al.(2019{\natexlab{a}})Yanaka, Mineshima, Bekki, Inui,
  Sekine, Abzianidze, and Bos}]{yanaka2019a}
Hitomi Yanaka, Koji Mineshima, Daisuke Bekki, Kentaro Inui, Satoshi Sekine,
  Lasha Abzianidze, and Johan Bos. 2019{\natexlab{a}}.
\newblock \href {https://doi.org/10.18653/v1/W19-4804} {Can neural networks
  understand monotonicity reasoning?}
\newblock In \emph{Proceedings of the 2019 ACL Workshop BlackboxNLP: Analyzing
  and Interpreting Neural Networks for NLP}, pages 31--40.

\bibitem[{Yanaka et~al.(2019{\natexlab{b}})Yanaka, Mineshima, Bekki, Inui,
  Sekine, Abzianidze, and Bos}]{yanaka2019b}
Hitomi Yanaka, Koji Mineshima, Daisuke Bekki, Kentaro Inui, Satoshi Sekine,
  Lasha Abzianidze, and Johan Bos. 2019{\natexlab{b}}.
\newblock \href {https://doi.org/10.18653/v1/S19-1027} {{HELP}: A dataset for
  identifying shortcomings of neural models in monotonicity reasoning}.
\newblock In \emph{Proceedings of the Eighth Joint Conference on Lexical and
  Computational Semantics (*{SEM} 2019)}, pages 250--255.

\end{thebibliography}


\end{document}